# OPTIMAL FUZZY MODEL CONSTRUCTION WITH STATISTICAL INFORMATION USING GENETIC ALGORITHM

Md. Amjad Hossain[1], Pintu Chandra Shill[2], Bishnu Sarker[1], and Kazuyuki Murase[2]

[1]Department of Computer science and Engineering, KUET, Khulna 9203, Bangladesh
`amjad_kuet@yahoo.com, bishnukuet@gmail.com`

[2]Department of System Design Engineering, University of Fukui, 3-9-1 Bunkyo, Fukui 910-8507, Japan
`pintu@synapse.his.u-fukui.ac.jp, murase@synapse.his.u-fukui.ac.jp`

## ABSTRACT

*Fuzzy rule based models have a capability to approximate any continuous function to any degree of accuracy on a compact domain. The majority of FLC design process relies on heuristic knowledge of experience operators. In order to make the design process automatic we present a genetic approach to learn fuzzy rules as well as membership function parameters. Moreover, several statistical information criteria such as the Akaike information criterion (AIC), the Bhansali-Downham information criterion (BDIC), and the Schwarz-Rissanen information criterion (SRIC) are used to construct optimal fuzzy models by reducing fuzzy rules. A genetic scheme is used to design Takagi-Sugeno-Kang (TSK) model for identification of the antecedent rule parameters and the identification of the consequent parameters. Computer simulations are presented confirming the performance of the constructed fuzzy logic controller.*

## 1. INTRODUCTION

**A** fuzzy rule-based model is a set of fuzzy IF-THEN rules that maps inputs to outputs. It has numerous practical applications in control [1], prediction and inference [2, 3] and has been found to be quite successful in examining problems with uncertainty and non-linearity. Fuzzy rules are either provided by human experts or learned from sample data. Many decision-making and problem-solving tasks are too complex to be understood quantitatively. However, people succeeded by using knowledge that is imprecise rather than precise. Since knowledge can be expressed in a more natural way by using fuzzy sets, many engineering and decision problems can be greatly simplified.

Most of fuzzy logic controllers (FLCs) to date have been static and based upon knowledge derived from imprecise heuristic knowledge of experienced operators. The construction of FLCs based on this type of expert knowledge can be quick and effective, provided the expert knowledge is available. On the other hand, without such an expert knowledge the design of FLCs can be slow as it relies on trial and error rather than a guided approach. So we need an automated knowledge acquisition method for FLCs which will be able to improve the overall performance in fuzzy control. For most fuzzy logic control problems, the most important issue is to determine the parameters that define the MFs. Because of this, the MFs optimization problems can be converted to parameter optimization problems. These parameters are generally based on the expert knowledge that is derived from heuristic knowledge of experienced control engineers and/or generated automatically. A variety of methods such as genetic algorithms (GAs), neural networks (NNs), self-organizing feature map (SOFM), tabu search (TS), and particle swarm optimization (PSO) have been used to improve the behavior of parameter optimization problem as well as selection and definition of fuzzy rules.

GA was used by Belarbi [4] in fuzzy rule base minimization. He applied GA to design FLC for the control of the pole and cart system and the control of the concentration in continuously stirred tank reactor. Arslan and Kaya [5] presented a method to adjust the shapes of MFs using GA. They have designed a fuzzy logic control system having single input and output. Meredith [6] has also applied GA to the fine tuning of MFs in a FLC for a helicopter. Bagis [7] have presented an approach for the determination of the structure and parameters of fuzzy RB. He applied this approach in the modeling of the nonlinear or complex systems. Bai and Chen [8] have proposed an automatic method for students' evaluation task. The purpose was to automatically construct the grade MFs of lenient-type grades, strick-type grades, and normal-type grades of fuzzy rules. Yang and Bose [9] have presented a method for generating fuzzy MFs with an unsupervised learning using SOFM. The SOFM approach is a two-step procedure. Firstly, it generates the proper clusters and secondly, it generates the fuzzy MFs according to the clusters in the first step. They have applied this method to pattern recognition. Evolutionary design of fuzzy logic based position controller for mobile robot is presented in [20]. Huang [10] has presented a fuzzy knowledge integration technique based on the PSO. His proposed approach consists of two phases: Firstly, it encodes the fuzzy rule sets and fuzzy sets with its corresponding MFs. Secondly; the particle swarm algorithm is used to explore the fuzzy rule sets, fuzzy sets and MFs to its optimal or the approximately optimal extent.

This paper proposes a new approach based on genetic algorithm (GA) for the optimum design of FLCs involving large number of parameters. The GA is employed as a self-adaptive learning strategy to automatic identification of the antecedent rule parameters and the identification of the consequent parameters is done by the least square method to design TSK fuzzy model.

The rest of the paper is organized as follows: Section II illustrates the literature review relevant to the FLCs. Statistical information criteria is describe in section III. Section IV describes how orthogonal transformation technique is used to order the importance of fuzzy rules. The inference mechanism of TSK type fuzzy logic controller is described in section V. In section VI we have introduced the key ideas of our proposed approach. This section also presents the methodology adopted for solving the nonlinear plant modeling problems in fuzzy environment. We describe our simulation and the results in Section VII and comparative analysis on nonlinear plant modeling problem and finally section VIII presents some concluding remarks based on the present study.

## 2. FUZZY LOGIC CONTROLLER

The idea of fuzzy logic was first introduced in 1960s by Professor Lofti Zadeh [11].

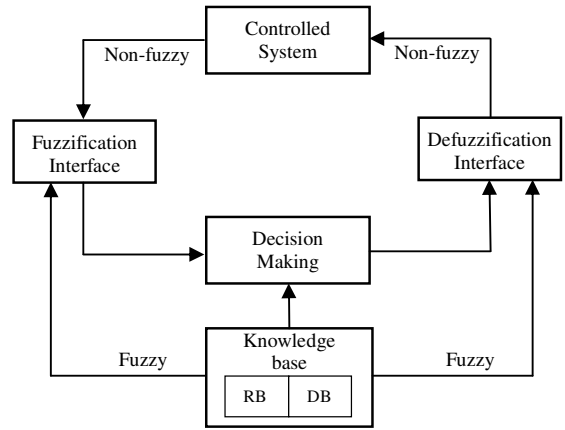

Figure 1. Fuzzy Logic Controller (FLC)

The general configuration of a FLC can be divided into four main parts; fuzzification interface, a rule base, an inference mechanism and defuzzification interface (Fig.1). The fuzzy system works as follows [12]: i). Determine the fuzzy membership values activated by the inputs. ii).Determine which rules are fired in the rule set. iii). Combine the membership values for each activated rule using the AND operator. iv). Trace rule activation membership values back through the appropriate output fuzzy membership functions. v). Utilize defuzzification to determine the value for each output variable. vi). Make decision according to the output values.

### 2.1. Fuzzifier

Fuzzy sets (Fig. 2) can be defined as $\mu_A$, membership function that associates with each element $x \in X$ where $X$ is the universe of discourse, a number called membership grade $\mu_A(x) \in [0,1]$. The *function* of the *fuzzifier* is to map a crisp input value $x \in X$ into a fuzzified value in $A \in U(universe)$. In this paper, we have used Non-singleton fuzzifier: $\mu_A(x_i)$ realizes maximum value 1 at $x = x_i$ and decrease from 1 to 0 while moving away from $x = x_i$.

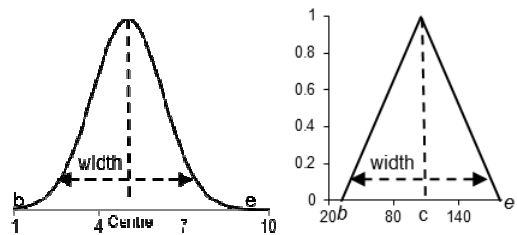

Figure 2. Fuzzy set (MFs: Gaussian and Triangular)

### 2.2. Fuzzy rule base

The general form of a fuzzy rule used in most FLCs is as follows:

$R^l$ : IF $x_1$ is $F_1^l$,........, and $x_n$ is $F_n^l$, THEN $y$ is $G^l$.

$l = 1, 2, 3,......., M$, $M$ = number of rules in the rule base

where $x_1, x_2,......., x_n$, and $y$ are the input and output linguistic variables respectively. $F_i^l$ and $G^l$ are fuzzy sets in input sets $X \in X_1 \times X_2 \times X_n$ and output sets $Y$. Each fuzzy IF-THEN rule has an antecedents (or IF) part containing several preconditions and a consequent (THEN) part describing the output action.

### 2.3. Fuzzy Inference Engine

A fuzzy relation $R^l$ can be defined as:

$R^l : X \times Y = \{(\vec{x}, y) : x \in X, y \in Y\}$ where $\vec{x}$ is a vector of the form $(x_1, x_2, .....x_n)^T$. This relation $R^l$ is the actual process of mapping from fuzzy sets in *X* to fuzzy sets in *Y*. $F_1^l \times F_2^l \times ....F_n^l \to G^l$, can be called fuzzy inference process. The process involves MFs, fuzzy logic operators, and if-then control rules. Fuzzy Inference process involves application of the fuzzy operators (AND or OR) in the antecedent, implication from the antecedent to the consequent, and aggregation of the consequents across the rules.

## 2.4. Defuzzifier

The final crisp output values are determined using a procedure known as "defuzzification process". The "Center of gravity" method is used as defuzzification method. Defuzzification produces a numerical (point-estimate) output value for the fuzzy set. The defuzzification method is centroid defuzzification which uses the fuzzy centroid $\theta$ as output:

$$\bar{\theta} = \frac{\sum_{j=1}^{p} \theta_j m_o(\theta_j)}{\sum_{j=1}^{p} m_o(\theta_j)}$$

where *O* defines a fuzzy subset of the universe of discourse $\tau = \{\theta_1, \theta_2 .....\theta_p\}$, *m* is the respective MF.

## 2.5. Membership functions(MFs)

A membership function is a curve that defines how each point in the input space is mapped to a membership value (or degree of membership) between 0 and 1. The membership function can be linear or nonlinear. Commonly used are left_triangle, right_triangle, triangle, and Gaussian [12] as shown in Fig. 3 [12].

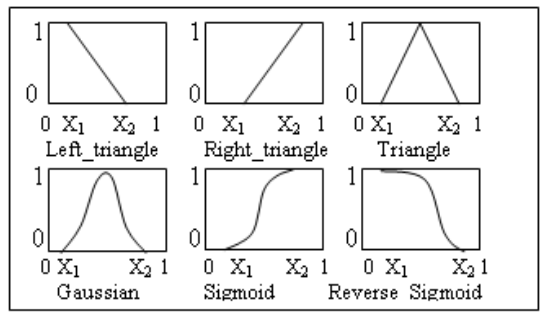

Figure 3. Membership Functions (MFs)

Left_triangle membership function:

$$f_{Left\_triangulae} = \begin{cases} 1, & \text{if } x < x_1 \\ \dfrac{x_2 - x}{x_2 - x_1}, & \text{if } x_1 \leq x \leq x_2 \\ 0, & \text{if } x > x_2 \end{cases} \quad (1)$$

Right_triangle membership function:

$$f_{Right\_triangulae} = \begin{cases} 1, & \text{if } x < x_1 \\ \frac{x - x_1}{x_2 - x_1}, & \text{if } x_1 \leq x \leq x_2 \\ 0, & \text{if } x > x_2 \end{cases} \quad (2)$$

Triangle membership function:

$$f_{triangulae}(x) = \begin{cases} 0, & \text{if } x < x_1 \\ 2\frac{x - x_1}{x_2 - x_1}, & \text{if } x_1 \leq x \leq \frac{x_2 + x_1}{2} \\ 2\frac{x_2 - x}{x_2 - x_1}, & \text{if } \frac{x_2 + x_1}{2} < x \leq x_2 \\ 1, & \text{if } x > x_2 \end{cases} \quad (3)$$

Gaussian membership function:

$$f_{Gaussian}(x) = e^{-0.5y^2} \quad \text{where } y = \frac{8(x - x_1)}{x_2 - x_1} - 4 \quad (4)$$

## 3. STATISTICAL INFORMATION CRITERIA

To estimate θ, the following criterion, known as AIC was designed by Akaike [5]

$$AIC(m) = -2\log f(y | \hat{\theta}) + 2m \quad (5)$$

Where y is the number observations, $\hat{\theta}$ is the reasonable estimate for θ based on y and *m* is the number of the unknown parameters, i.e., the dimension of θ. The following equation is the equivalent one of (5) providing superior convenience to compute:

$$AIC(m) = \log(\hat{\sigma}_\varepsilon^2) + \frac{2m}{N} \quad (6)$$

Where, $\hat{\sigma}_\varepsilon^2$ is the estimated variance of model residuals. However, we will refer to this criterion as the BDIC in the remaining sections of this paper, that is,

$$BDIC(m) = \log(\hat{\sigma}_\varepsilon^2) + \frac{\alpha m}{N} \quad (7)$$

On the other hand, we will have another criterion which will be referred to as SRIC for simplicity:

$$SRIC(m) = \log(\hat{\sigma}_\varepsilon^2) + \frac{\log(N)m}{N} \quad (8)$$

It was independently derived by Schwarz and Rissanen, in which they used log(N) to replace the constant "2" in (6). There are some other criteria that are similar to AIC having a common feature such as they all pursue a balance between the goodness of fit and the model complexity. Moreover, all these information criteria can be examined under a general framework,

$$IC(m) = \log(\hat{\sigma}_\varepsilon^2) + mp(N) \quad (9)$$

Where *p (.)* is a positive function of the sample size and satisfies the condition

$$Lim_{N \to \infty} P(N) = 0 \quad (10)$$

## 4. ORDERING FUZZY RULES USING ORTHOGONAL TRANSFORMATION TECHNIQUE

For ordering fuzzy rules, here we describe an approach based on a numerically reliable orthogonal transformation technique which is a fuzzy model with constant consequent constituents. For simplicity, we call this type of model as constant fuzzy model which has the following form [5]:

$$R_i : \text{If } x_1 \text{ is } A_{i1} \text{ and........and } x_p \text{ is } A_{ip}$$
$$\text{then } y = c_i, i = 1,2,...,m_r$$

Where $p$ and $m_r$ are the number of input variables and rules respectively. $x_j$, $j = 1, 2, \cdots p$ is the $j^{th}$ input variable and $A_{ij}$ are the membership functions of input variables, $c_i$ are constant consequent constituents. y is the total output variable of the model. If the firing strength of the $i^{th}$ rule in the model is

$$w_i = A_{i1}(x_1) \times A_{i2}(x_2) \times ..................A_{ip}(x_p)$$

Then the total output of the model is computed as follows:

$$y = \frac{\sum_{i=1}^{m_r} w_i c_i}{\sum_{i=1}^{m_r} w_i} = \sum_{i=1}^{m_r} v_i c_i \tag{11}$$

Where, $v_i$ is the normalized firing strength of the $i^{th}$ rule defined by

$$v_i = \frac{w_i}{\sum_{i=1}^{m_r} w_i} \tag{12}$$

## 5. TAKAGI-SUGENO-KANG (TSK) FUZZY MODEL

The TSK fuzzy model [5] consists of IF-THEN rules where the rule consequents are usually constant values (singletons) or linear functions of the inputs.

$$R_i = IF \ x_1 \text{ is } A_{i1} \text{ and.......} x_n \text{ is } A_{in}, \text{ Then } y_i = c_{io} +$$
$$c_{i1}x_1 + ........... + c_{in}x_n \text{ For i} = 1,2,3,......, N_R$$

Where $N_R$ is the number of rules, $x = [x_1, x_2,................,x_n]$ is the input vector, $y_i$ is the output of the $i^{th}$ rule, $A_{ij}$ are the antecedent fuzzy sets that are characterized by membership functions (MFs) $\mu_{A_{ij}}(x_j)$, and $c_{ij}$ are real-valued weight parameters. The model output is computed by

$$y = \frac{\sum_{i=1}^{N_R} \tau_i y_i}{\sum_{i=1}^{N_R} \tau_i} = \frac{\sum_{i=1}^{N_R} \tau_i (c_{i0} + c_{i1}x_1 + ..........+ c_{in}x_n)}{\sum_{i=1}^{N_R} \tau_i},$$

where $\tau_i$ is the firing strength of the rule $R_i$ which is defined as

$$\tau_i = A_{i1}(x_1) \times A_{i2}(x_2) \times ...............\times A_{in}(x_n) \tag{13}$$

## 6. INTEGRATED MODEL AND GENETIC DESIGN PROCESS

The integrated architecture depicted in Fig. 4 is a fuzzy system that uses GA to determine fuzzy sets and fuzzy control rules.

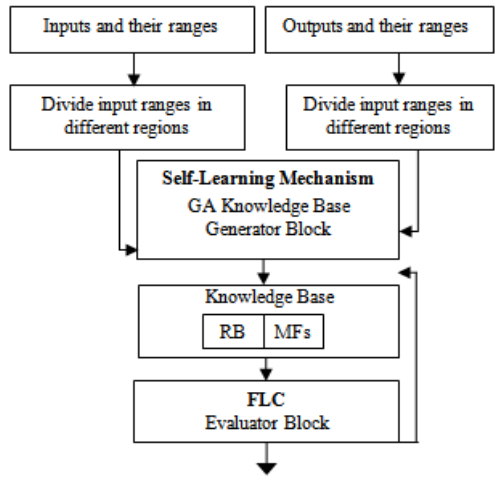

Figure 4. Integration of type-2 FLCs and GA

In this paper, we employed GA to optimize the parameters of the MFs of FLC; we consider using Gaussian Interval MFs to input variables. At the same time, we also employed GA for the selection and definition of rule base of FLC.

### 6.1. Representation

When designing a fuzzy model using a genetic algorithm, the most important considerations are the chromosome representation scheme, that is, how to encode the fuzzy system into the chromosome and how to evaluate the potential solution. The objective of the genetic optimization is to find an optimal value of centers and widths of the membership functions.

### 6.2. Evaluation

Defining a proper fitness function is one of the most important issues when using a genetic approach. The most common way to define a fitness function is to measure the performance of an individual in terms of the mean-squared-error (MSE) [14]:

$$MSE(S_k) = \frac{1}{N_T} \sum_{h}^{N_T} (y_h - \hat{y}_h)^2 \quad (14)$$

where $y_h$ is the $h^{th}$ desired output and $\hat{y}_h$ is the $h^{th}$ model output.

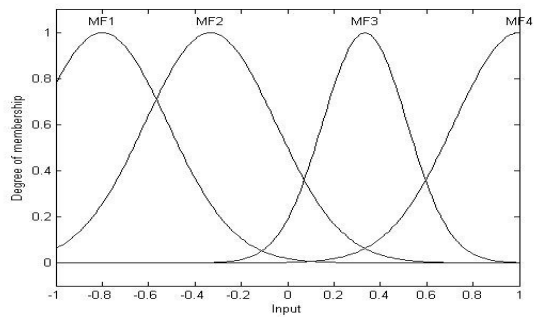

Figure 5. Definition of the overlapping length in the penalty function.

The penalty function actively calculates the degree of overlapping between two MFs and is defined as follows:

$$PF(s_k) = \sum_{j=1}^{N_I} \frac{\sum_{i=1}^{} \lambda_{ij}}{|\chi_j|} \quad (15)$$

where $\lambda_{ij}$ is the length of the $i^{th}$ overlapping occurrence between two MFs in the $j^{th}$ input domain. $\chi_j$ is the length of the $j^{th}$ input domain. The specific level that constrains the overlapping between two MFs is denoted by ξ as shown in Fig. 5. Using this penalty value with the MSE value, the fitness function is defined as follows:

$$F(S_k) = \frac{1}{MSE(s_k) + \beta.PF(s_k)} \qquad (16)$$

Where $\beta$ is a design parameter that is used to make a compromise between the MSE and the penalty function.

### 6.3. Genetic Operators

#### 6.3.1. Reproduction

According to the fitness value of each individual, we first apply a ranking method. After ranking all the individuals in the population according to their fitness value, the upper 30% of the population is used to generate 50% of the new population [14].

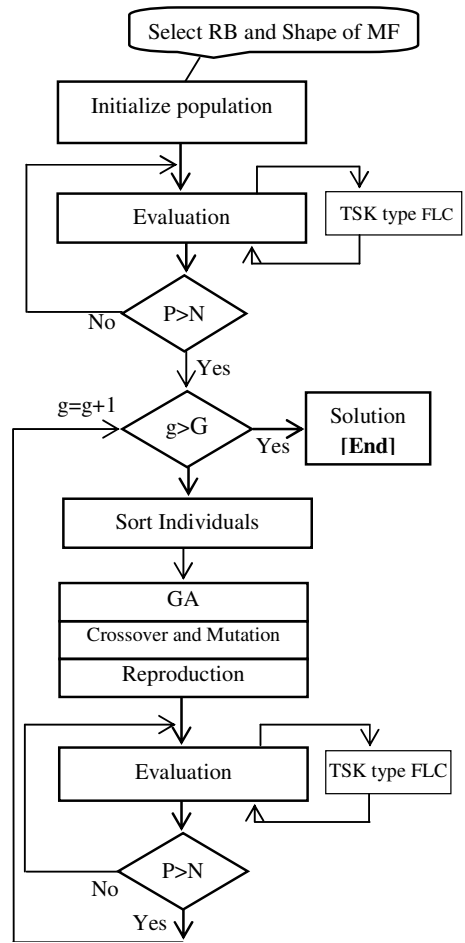

Figure 6. Combined TSK type FLC and GA Algorithm

### 6.3.2. Crossover

Crossover is the process of exchanging portions of two 'parent' individuals. An overall probability is assigned to the crossover process, which is the probability that given two parents, the crossover operation will occur. For convenience, we rewrite (13) as $S_k(t) = [p_1, p_2............p_L]$ where $p_i$ corresponds to $A_{ij}$ in (13). We have used two types of crossover operations in this paper. The first is a bitwise crossover. When two parents $s_v(t)$ and $s_w(t)$ are selected for the crossover operation, changing point $t$ is selected randomly within the range of an individual and swapping occurs as follows:

$$S_v(t+1) = [v_1 v_2..........v_k w_{k+1}........w_L], \ S_w(t+1) = [w_1 w_2..........w_k v_{k+1}........v_L$$

The next crossover is an arithmetic crossover operator, which produces children using a linear combination of two parents as follows:

$$S_v(t+1) = \alpha.s_v(t) + (1-\alpha).s_w(t), \ S_w(t+1) = \alpha.s_w(t) + (1-\alpha).s_v(t)$$

where the parameter α is generated randomly each time the arithmetic crossover is applied.

### 6.3.3. Mutation

Mutation consists of changing an element's individual value at random, often with a constant probability. Mutation is performed column-wise for every center value and every width value of all individuals as follows: $\theta(t+1) = \theta(t) + N(0, \rho)$, where $\theta$ is a parameter value: the center or the width of a membership function and $N(0, \rho)$ represents normal distribution function.

### 6.4. Genetic Procedure

The combined fuzzy logic controller and genetic algorithm is shown in Fig. 6. The framework of genetic procedure used in this paper is as follows [15]:

**Step 1.** Generate an initial population P (0) = $[s_1(0) s_2(0)…………….....s_{N_P}(0)]$ at random and set $i$ =0.
**Step 2.** Using the parameter values of each individual, constructs a fuzzy model and calculates consequent parameters for all individuals.
**Step 3.** Evaluate every individual.
**Step 4.** Apply genetic operators to obtain the next (population $Pi$ +1).
**Step 5.** $i=i+1$, return to step 2) if the $G_{max}$ is not reached or the procedure is terminated.

## 7. IMPLEMENTATION AND RESULTS

In this section we present the simulation results of our proposed algorithm using different configurations. The implementation of the fuzzy model is written in c++ and complied using the Borland c++ 4.5 compiler. Here we have implemented constant fuzzy model, TSK type fuzzy model and membership constrained based TSK model. We have also presented a comparative analysis among these three types of fuzzy model. We have implemented membership constrained based TSK type fuzzy model by using genetic algorithm with a proper fitness function which have been discussed previously. We have also introduced the orthogonal transformation based technique for determining the importance of fuzzy rules generated through genetic algorithm. The proposed optimality criteria can be applied for the construction of optimal fuzzy models. In this section, first of all a nonlinear plant modeling problem is described and then the results of applying the proposed optimality criteria to construct two types of fuzzy model is discussed.

## 7.1. Nonlinear System Identification Second Order System

We have used the following Nonlinear System Identification second Order System to illustrate the application of the optimality criteria in the construction of optimal fuzzy models [13]. The plant to be identified is described by the following second order difference equation:

$$Y(k) = f(y(k-1), y(k-2)) + u(k)$$

Where, $$f(y(k-1), y(k-2)) = \frac{y(k-1)y(k-2)[y(k-1) - 0.5]}{1 + y(k-1)^2 + y(k-2)^2}$$

Here, *f* is the nonlinear component of the plant which is usually called the "unforced system" in control literature. It has an equilibrium state *(0, 0)* in the state space. Fig. 7 shows the trajectory of the unforced system in the state space.

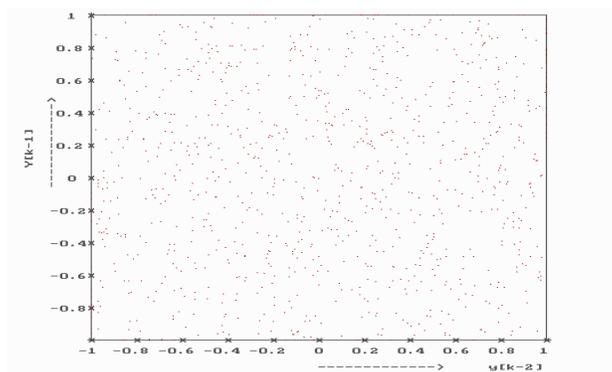

Figure 7. Trajectory of nonlinear plant modeling equation

The main purpose of this research is to estimate *f* using a fuzzy model. For this approximation a simulated data set of 1200 items is generated from the above plant model. The first 1000 data elements are obtained by assuming a random input signal u(k) uniformly distributed in the interval [-1,1] and this data set is used to build a fuzzy model. To test the performance of the resulting model other 200 data elements are generated by using a sinusoid input signal. $y(k-1)$ and $y(k-2)$ are selected as the input variables. The complete combination of membership functions constructed using a six-dimensional cubic B-splines for the two input variables generate 36 fuzzy rules and correspondingly, a 1000×36 firing strength matrix *p* can be constructed. These rules are labeled using the numbers 1, 2,···, 36 indicating the corresponding position of the rules in the rule base associated with a combination of membership functions. In particular, the number "1" denotes the first rule in the rule base associated with the membership functions combination $\{A_{11}(x_1), A_{21}(x_2)\}$ (where $x_1 = y(k-1), x_2 = y(k-2)$), while the number "36" indicates the 36$^{th}$ rule in the rule base associated with the membership functions combination $\{A_{16}(x_1), A_{26}(x_2)\}$.

## 7.2. Construction of an optimal fuzzy Model

As we can see, the Table 1 shows the importance order of fuzzy rules in the rule base.

Table 1. Order of Importance of Fuzzy Rules in the Rule Base.

| Importance Order | Rule position | Importance Order | Rule position | Importance order | Rule position |
|---|---|---|---|---|---|
| 1 | 16 | 13 | 27 | 25 | 1 |
| 2 | 21 | 14 | 26 | 26 | 36 |
| 3 | 22 | 15 | 29 | 27 | 34 |
| 4 | 15 | 16 | 11 | 28 | 33 |
| 5 | 20 | 17 | 7 | 29 | 32 |
| 6 | 8 | 18 | 2 | 30 | 31 |
| 7 | 10 | 19 | 4 | 31 | 25 |
| 8 | 23 | 20 | 35 | 32 | 24 |
| 9 | 9 | 21 | 30 | 33 | 19 |
| 10 | 28 | 22 | 13 | 34 | 18 |
| 11 | 14 | 23 | 5 | 35 | 12 |
| 12 | 17 | 24 | 3 | 36 | 6 |

Using this order, a constant fuzzy model with $m_r$ rules can be constructed. The rule whose singular value is greater is the most important rule which is shown in Fig. 8. A total of 36 models with different number of fuzzy rules (1 to 36) can be constructed this way. The Fig. 9 shows the values of information criteria which are obtained by applying the proposed optimality criteria AIC, BDIC, and SRIC to these models.

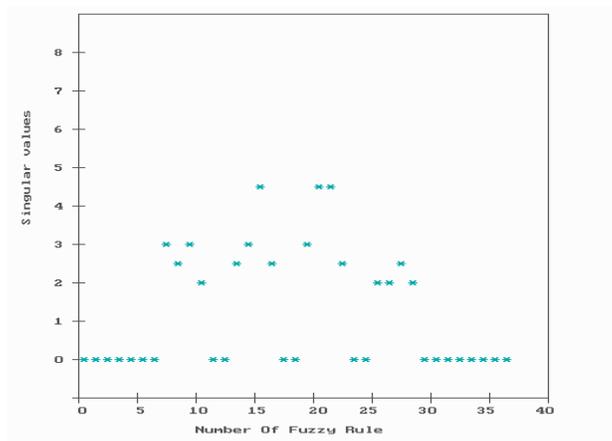

Figure 8. Distribution of singular values

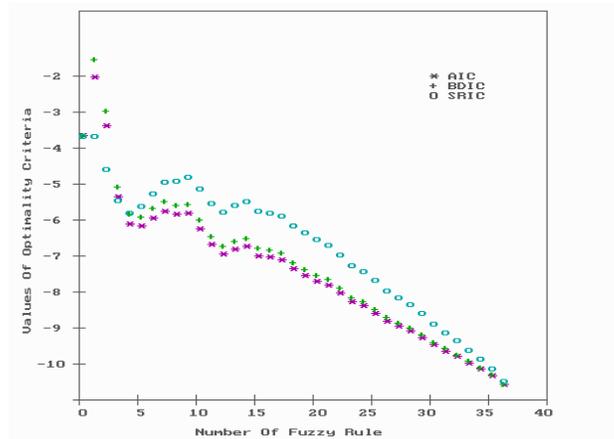

Figure 9. Values of information criteria for constant fuzzy models with varying number of fuzzy rules

These three criteria find different optimal models. It is observed that for both AIC and BDIC, the model with 36 rules gives a minimum value. On the contrary, the model with only 12 rules has the lowest SRIC value provided that the optimal rules are chosen between 1 and 15.

Table II shows the MSE's of the three constant models varying with number of fuzzy rules including the model consisting of the complete 36 rules for the training stage. It can be clearly seen that the model with highest number of rules (i.e. 36 rules) has the smallest MSE value. Trajectory of constant fuzzy model with 36 rules is shown in fig. 10.

TABLE II
PERFORMANCE OF THE CONSTANT FUZZY MODEL IN THE TRAINING STAGE

| Constant Fuzzy model | MSE |
| --- | --- |
| Constant model with 36 rules | 0.193603 |
| Constant model with 28 rules | 0.323083 |
| Constant model with 23rules | 0.418989 |

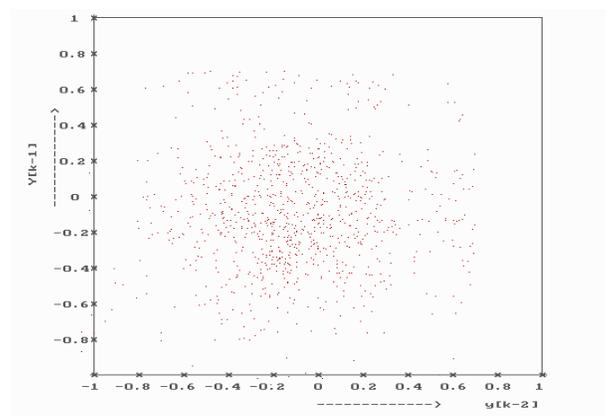

Figure 10. Trajectory of constant fuzzy model with fuzzy models with varying number of fuzzy 36 rules.

### 7.3. Construction of an Optimal TSK model

Similarly, a total of 36 TSK models can be constructed with varying number of fuzzy rules by selecting the $m_r$ most important rules. Applying AIC, BDIC, SRIC to these models, we obtain the result shown in Fig. 11.

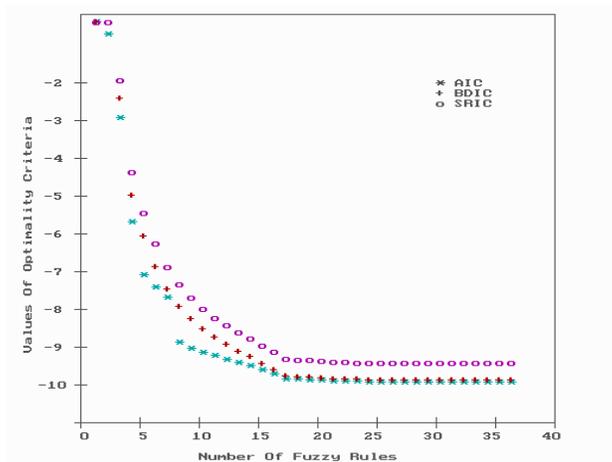

Figure 11. Values of information criteria for TSK model with varying number of fuzzy rules

TABLE III
PERFORMANCE OF THE TSK MODEL IN THE TRAINING STAGE

| TSK Fuzzy model | MSE |
| --- | --- |
| TSK model with 36 rules | 0.158978 |
| TSK model with 28 rules | 0.158978 |
| TSK model with 24rules | 0.156886 |

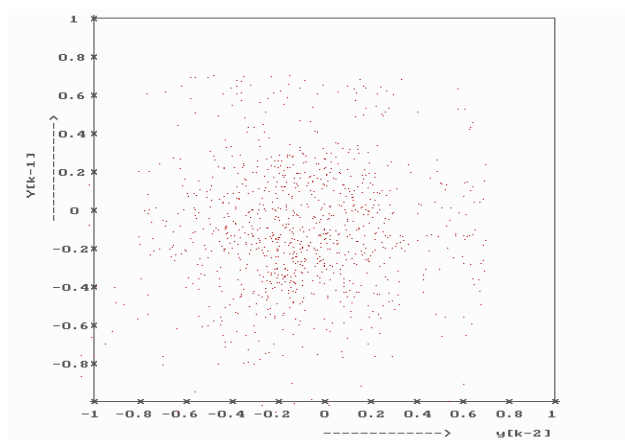

Figure 12. Trajectory of TSK model with varying number of fuzzy rules

Though the model with 24 rules yields the lowest AIC, BDIC and SRIC values but AIC has suggested a slightly over-fitted model. Table-III shows the MSE's of the two TSK models with 24 and 28 fuzzy rules as well as that of the TSK model consisting of the complete 36 rules.

It is observed that the 24-rule TSK model gives a better performance for fitting the training data than the 28-rule and 36-rule TSK model. Fig. 12 shows the trajectory of the optimal TSK model with 26 rules in the state space. Comparing Fig. 12 with Fig. 7, it can be seen that the trajectories of the identified TSK model and the real system are actually indistinguishable. Fig. 13 shows the outputs of the real plant and the optimal model.

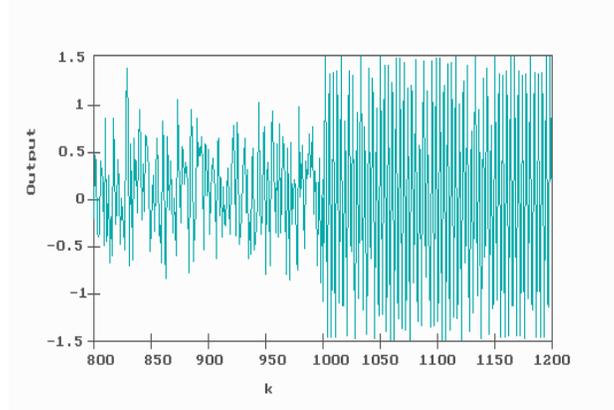

Figure 13. Output of the plant and the TSK model with 24 rules.

### 7.4. Comparison of Constant Fuzzy Models and TSK models

In this section we compare the performance of the constant fuzzy model and that of the TSK model. Fig. 14 compares these two models with respect to the logarithmic value of MSE's for different number of fuzzy rules.

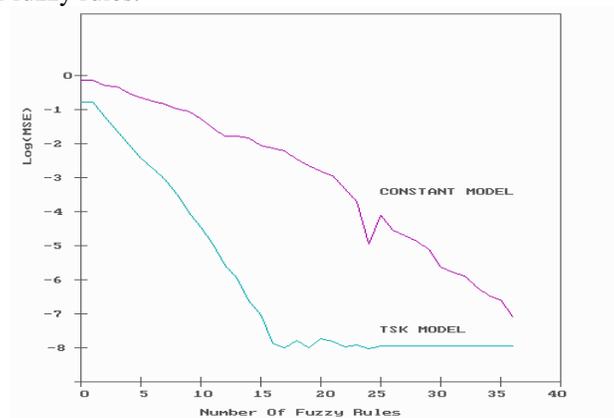

Figure 14. Comparison of constant fuzzy model and TSK model with varying number of fuzzy rules

In order to achieving the same accuracy level the TSK model needs far less rules than the constant fuzzy model. This finding is consistent since each rule in TSK model contains more important information than that in the constant fuzzy model. From the Fig. 14 we have shown that TSK model is more optimal than Constant fuzzy model.

### 7.5. Construction of TSK Model Using Genetic Algorithm

Using genetic algorithm we determine the range of antecedent and consequent parameters with a proper fitness function which is described in previous section. Then we construct TSK model by calculating weight parameters. Some matrix manipulations are used for calculating weight parameters.

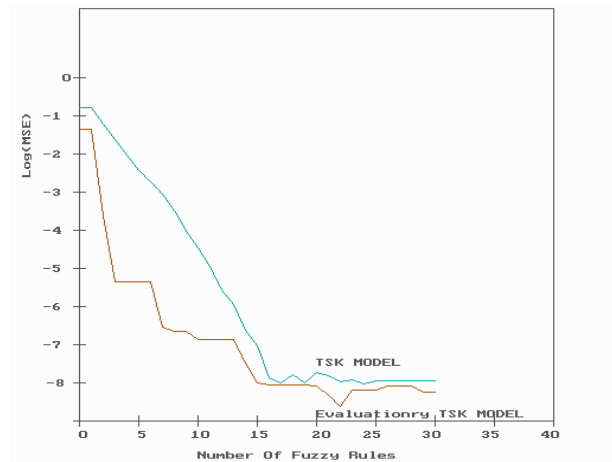

Figure 15. Comparison between rule-based TSK model and membership constrained based (evolutionary) TSK model.

In this algorithm we used 100 generations, 60 individuals and Gaussian membership function. This membership constrained based TSK model is compared with the rule based TSK model and it is shown in Fig.15. We can see that the TSK model designed by genetic algorithm based on membership functions is more optimal than rule based TSK model.

## 8. CONCLUSIONS

In this work, a genetic optimization approach is proposed to identify the optimal fuzzy model as well as fuzzy control rules and MFs. Since the complexity of a fuzzy model is determined not only by the number of antecedent and consequent parameters but also by the number of fuzzy rules, so we use three statistical information criterions such as AIC, BDIC and SRIC to reduce fuzzy rules . The simulation results suggest that SRIC is likely to be a better optimality criterion for fuzzy model identification. Three types of fuzzy model such as constant, rule based TSK and membership constrained based TSK model is implemented and compared their performance with respect to the logarithmic value of mean square error (MSE) which is the fitness function. We have shown that the TSK model performs better than constant fuzzy model. Finally, it is seen that the membership constrained based TSK model, implemented by genetic algorithm outperforms all other fuzzy models.

International Journal of Computer Science & Information Technology (IJCSIT) Vol 3, No 6, Dec 2011

**Authors**

**Md. Amjad Hossain** received the B.Sc. Engg. degree (with honors) in Computer Science and Engineering (CSE) from Khulna University of Engineering & Technology(KUET), Khulna - 9203, Bangladesh, 2008. He is currently serving as a Lecturer in Department of CSE, KUET. He has published several research papers in some reputed Journal and Conference. His current research interests include evolutionary computation, cloud computing, mobile computing and fuzzy Logic.

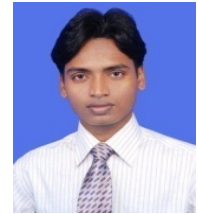

**Pintu Chandra Shill** received the B.Sc. degree in Computer Science Engineering (CSE) from Khulna University of Engineering and Technology (KUET), Bangladesh in 2003 and M.Sc. degree in Computer Engineering from Politecnico di Milano, Italy in 2008. He joined as a lecturer at the Department of CSE, KUET in 2004 and currently he is serving as an Assistant Professor. He has published several research papers in some reputed Journal and Conference. His research interest includes evolutionary computation, fuzzy logic and artificial neural networks.

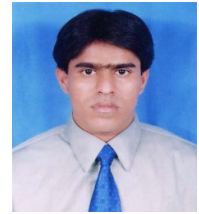

**Bishnu Sarker** received the B.Sc. degree in Computer Science Engineering (CSE) from Khulna University of Engineering and Technology (KUET), Bangladesh in 2011. He joined as a lecturer at the Department of CSE, KUET in 2011 and currently he is serving as a Lecturer. His research interest includes Evolutionary Computation, Fuzzy Logic, Artificial Neural Networks, Distributed Computing, and Fault Tolerant Computing.

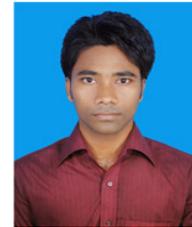

**Kazuyuki Murase** is a Professor at the Department of Human and Artificial Intelligence Systems, Graduate School of Engineering, University of Fukui, Fukui, Japan, since 1999. He received ME in Electrical Engineering from Nagoya University in 1978, PhD in Biomedical Engineering from Iowa State University in 1983. He Joined as a Research Associate at Department of Information Science of Toyohashi University of Technology in 1984, as an Associate Professor at the Department of Information Science of Fukui University in 1988, and became the professor in 1992. He is a member of The Institute of Electronics, Information and Communication Engineers (IEICE), The Japanese Society for Medical and Biological Engineering (JSMBE), The Japan Neuroscience Society (JSN), The International Neural Network Society (INNS), and The Society for Neuroscience (SFN). He serves as a Board of Directors in Japan Neural Network Society (JNNS), a Councilor of Physiological Society of Japan (PSJ) and a Councilor of Japanese Association for the Study of Pain (JASP).

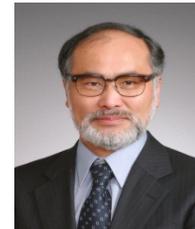